\documentclass[10pt,conference,a4paper]{IEEEtran}
\IEEEoverridecommandlockouts
\usepackage{cite}
\usepackage{amsmath,amssymb,amsfonts}
\usepackage{algorithm}
\usepackage{algpseudocode}
\usepackage{graphicx}
\usepackage{tabularx}
\usepackage{url}

\newcommand\blfootnote[1]{%
	\begingroup
	\renewcommand\thefootnote{}\footnote{#1}%
	\addtocounter{footnote}{-1}%
	\endgroup
}

\usepackage{textcomp}
\def\BibTeX{{\rm B\kern-.05em{\sc i\kern-.025em b}\kern-.08em
    T\kern-.1667em\lower.7ex\hbox{E}\kern-.125emX}}
\begin{document}

\title{Document Image Classification with Intra-Domain Transfer Learning and Stacked Generalization of \\Deep Convolutional Neural Networks\\
}

\author{
\IEEEauthorblockN{Arindam Das}
\IEEEauthorblockA{\textit{CDA DVS}\\
%\textit{Deep Learning for ADAS} \\
\textit{Valeo India Pvt. Ltd.}\\
Chennai, India \\
arindam.das@valeo.com}
\and
\IEEEauthorblockN{Saikat Roy}
\IEEEauthorblockA{\textit{Institute for Informatics} \\
\textit{University of Bonn}\\
Bonn, Germany \\
saikatroy@uni-bonn.de}
\and
\IEEEauthorblockN{Ujjwal Bhattacharya}
\IEEEauthorblockA{\textit{CVPR Unit} \\
\textit{Indian Statistical Institute}\\
Kolkata, India \\
ujjwal@isical.ac.in}
\and
\IEEEauthorblockN{Swapan K. Parui}
\IEEEauthorblockA{\textit{CVPR Unit} \\
\textit{Indian Statistical Institute}\\
Kolkata, India \\
swapan@isical.ac.in}

%\and

}

\maketitle

\begin{abstract}
In this article, a region-based Deep Convolutional Neural Network framework is presented for document structure learning. The contribution of this work involves efficient training of region based classifiers and effective ensembling for document image classification. A primary level of `inter-domain' transfer learning is used by exporting weights from a pre-trained \textit{VGG16} architecture on the ImageNet dataset to train a document classifier on whole document images. Exploiting the nature of region based influence modelling, a secondary level of `intra-domain' transfer learning is used for rapid training of deep learning models for image segments. Finally, a stacked generalization based ensembling is utilized for combining the predictions of the base deep neural network models. The proposed method achieves state-of-the-art accuracy of 92.21\% on the popular RVL-CDIP document image dataset, exceeding the benchmarks set by the existing algorithms.

\blfootnote{Preprint Copy. Accepted in \textit{24th International Conference in Pattern Recognition (ICPR), Beijing, China, 2018.}}

\end{abstract}


\section{Introduction}
Documents can be classified into various classes based on their text contents and/or their structural properties. During a manual search for a particular document from a large collection of documents, knowledge about the type or structure of the document helps reduce the time necessary for the search. However, the automatic accomplishment of the same is a challenging task. In an early study \cite{tang1991}, it was observed that a real-life document can be viewed in different ways, in both geometric and logical structure spaces. The authors observed that effective understanding of the document structure can be realized through the use of an expert system and pattern classification methods.

Automatic classification of document images is an effective initial step of various Document Image Processing (DIP) tasks such as document retrieval, information extraction and text recognition, among others. The performance of a DIP system may be enhanced through efficient initial classification of an input document into a number of pre-determined categories. Automatic classification also has a significant role in indexing the documents of a Digital Library. It has been seen that any large volume of documents from different categories can be better organized provided these are first classified into several categories based on their structures \cite{shin2001}.

 %\hyperref[https://doi.org/10.1016/j.patrec.2017.03.004]{link} }
%\blfootnote {DOI: https://doi.org/10.1016/j.patrec.2017.03.004}

\begin{figure*}[t]
	%\vspace*{-3mm}
	\begin{center}
		\includegraphics[height = 0.22\columnwidth]{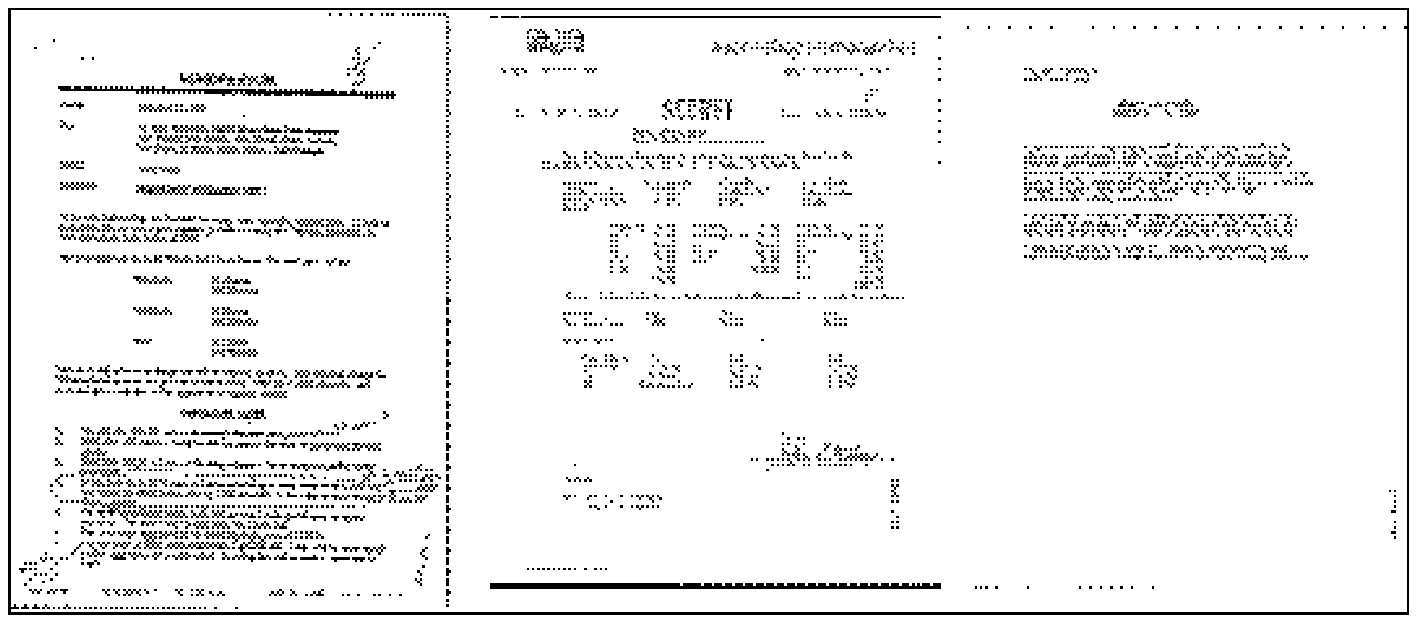}
		\hfill  \includegraphics[height = 0.22\columnwidth]{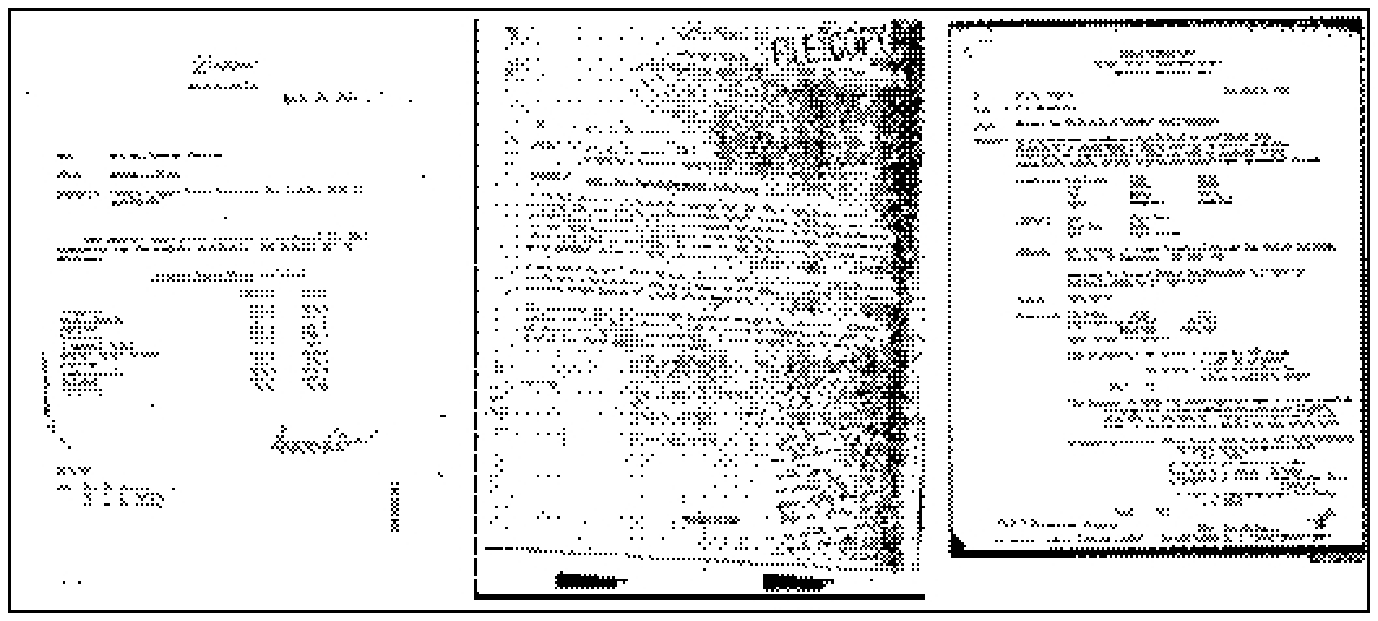}
		\hfill  \includegraphics[height = 0.22\columnwidth]{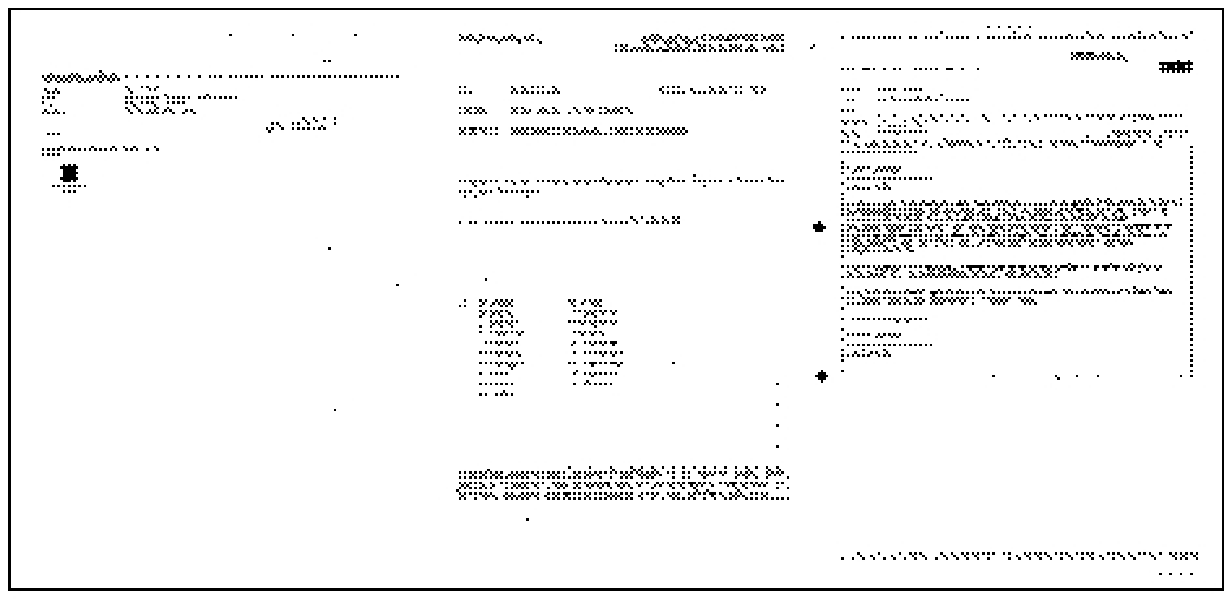}
		\hfill  \includegraphics[height = 0.22\columnwidth]{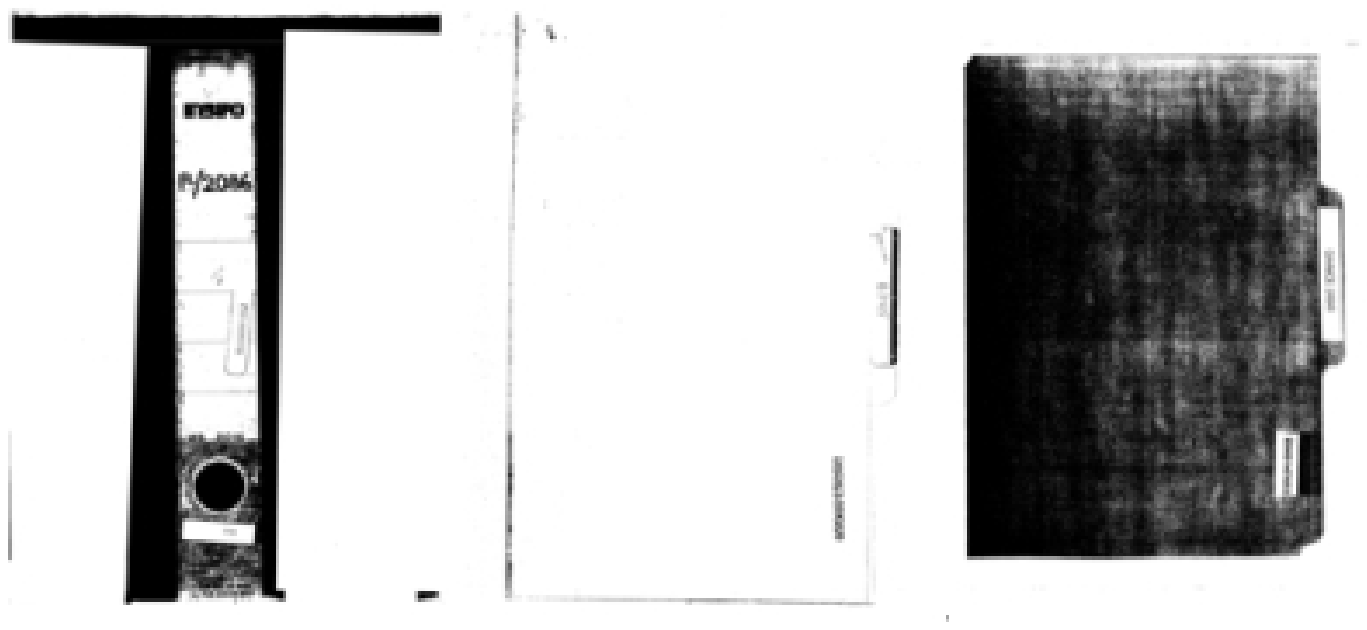}\\
		$(a)$\hspace*{4cm}$(b)$\hspace*{4cm}$(c)$\hspace*{4cm}$(d)$\\
		\vspace*{.25cm}
		\includegraphics[height = 0.22\columnwidth]{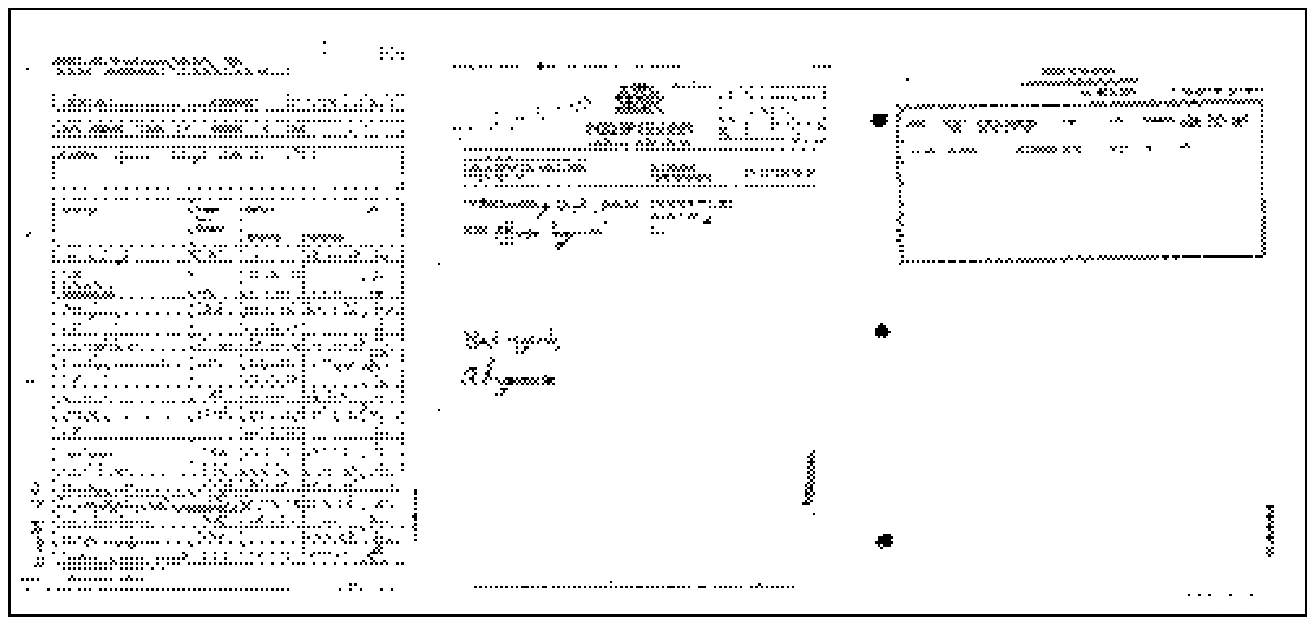}
		\hfill  \includegraphics[height = 0.22\columnwidth]{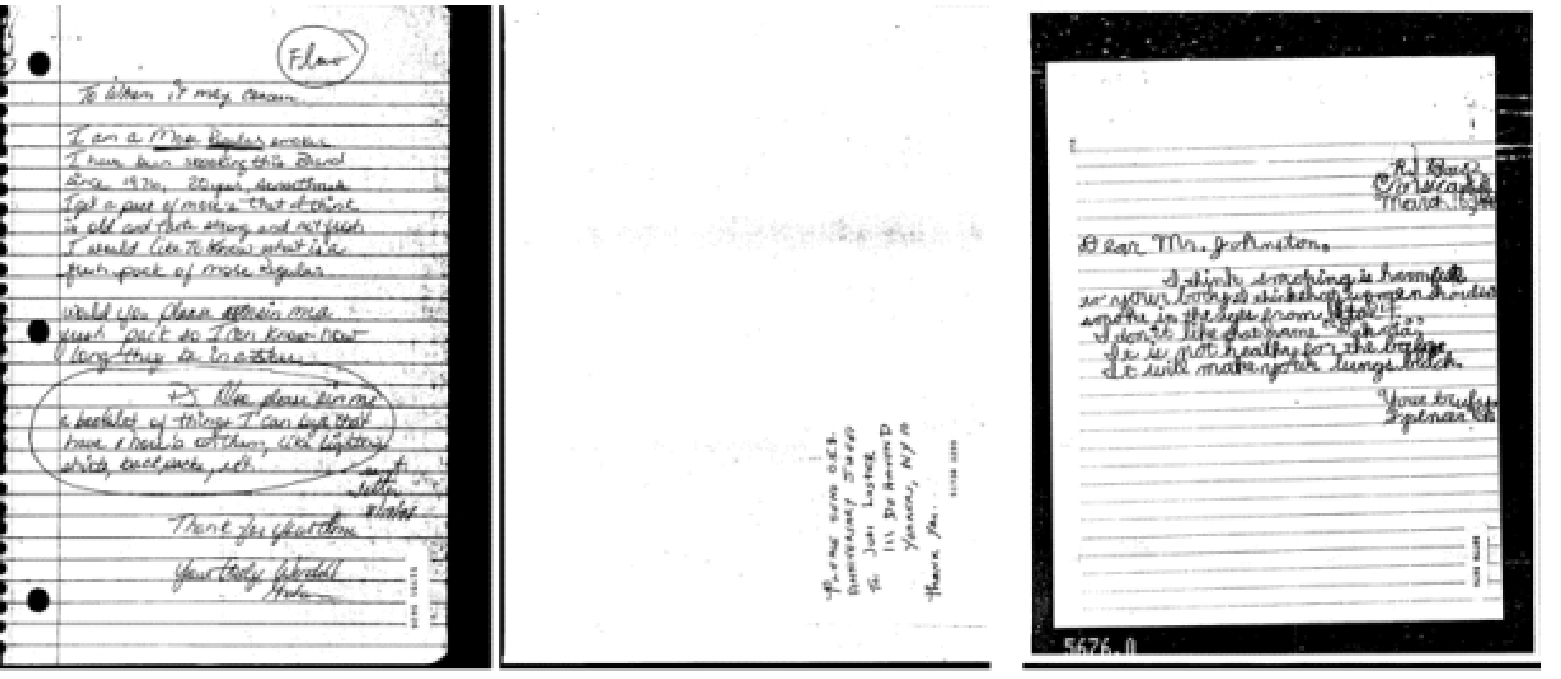}
		\hfill  \includegraphics[height = 0.22\columnwidth]{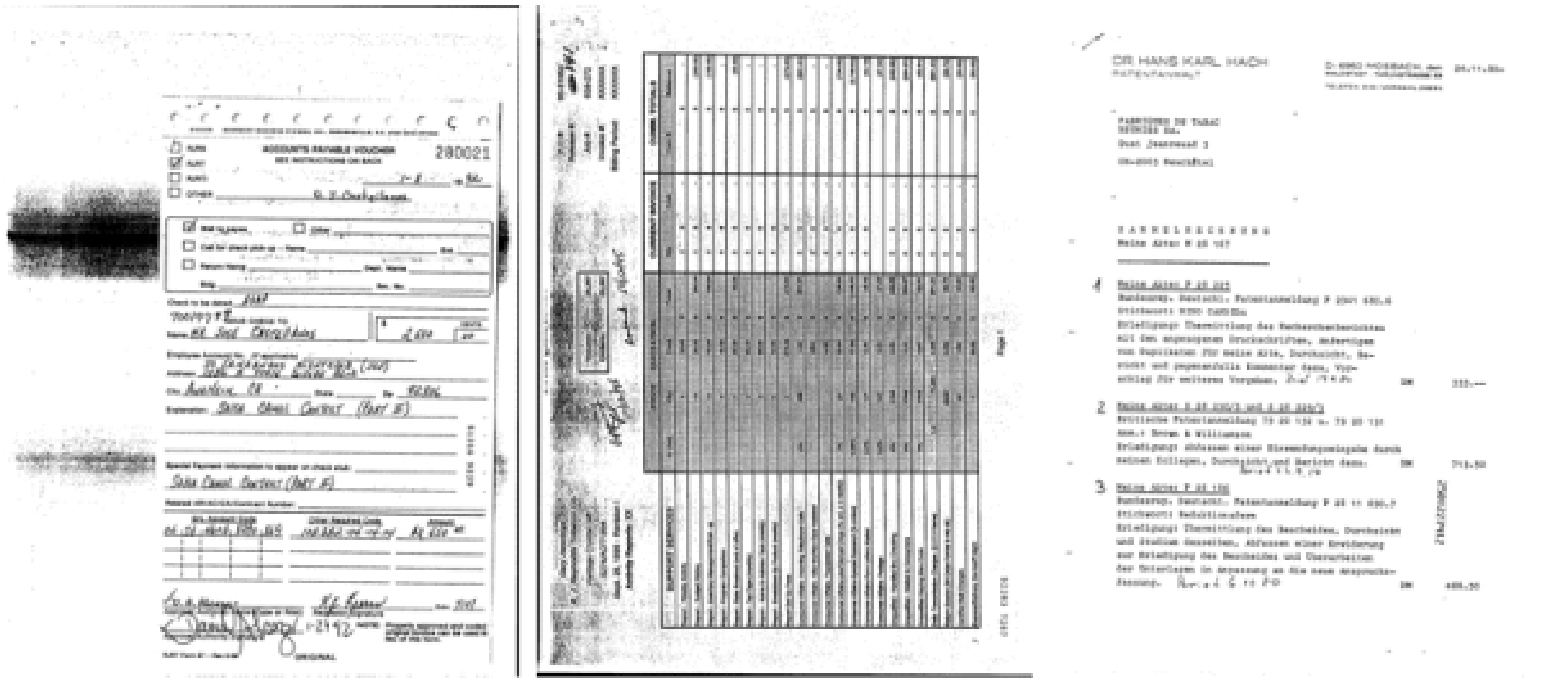}
		\hfill  \includegraphics[height = 0.22\columnwidth]{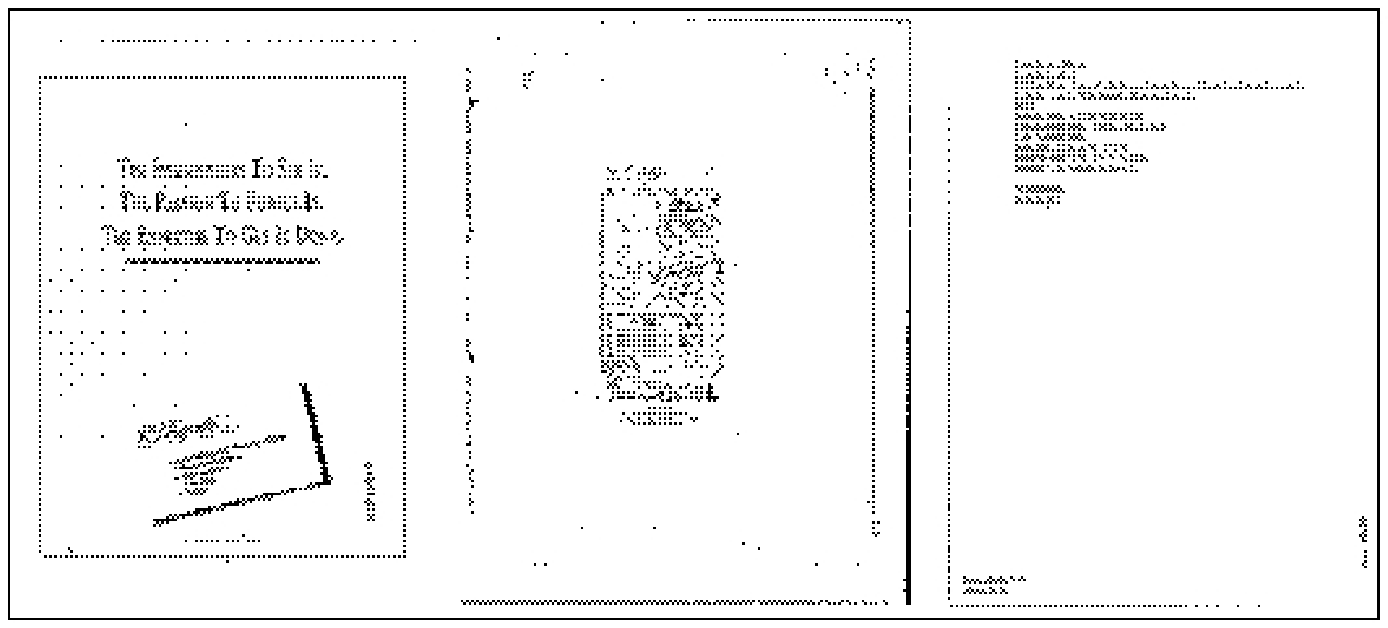}\\
		$(e)$\hspace*{4cm}$(f)$\hspace*{4cm}$(g)$\hspace*{4cm}$(h)$\\
		\vspace*{.25cm}
		\includegraphics[height = 0.22\columnwidth]{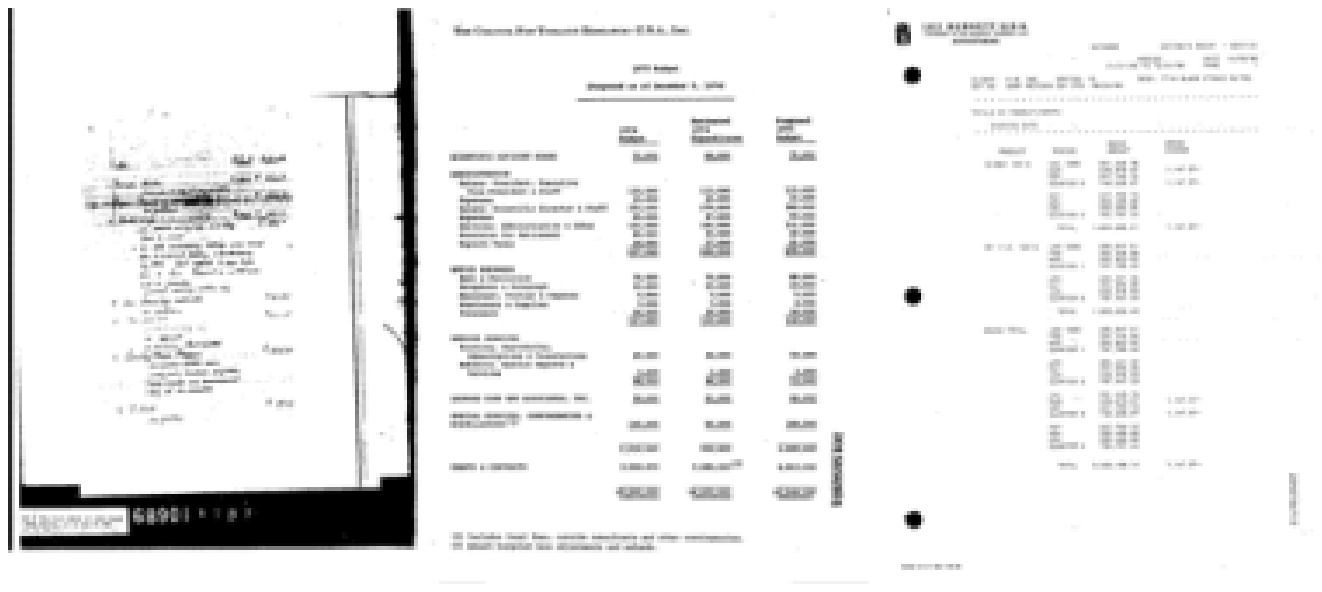}
		\hfill  \includegraphics[height = 0.22\columnwidth]{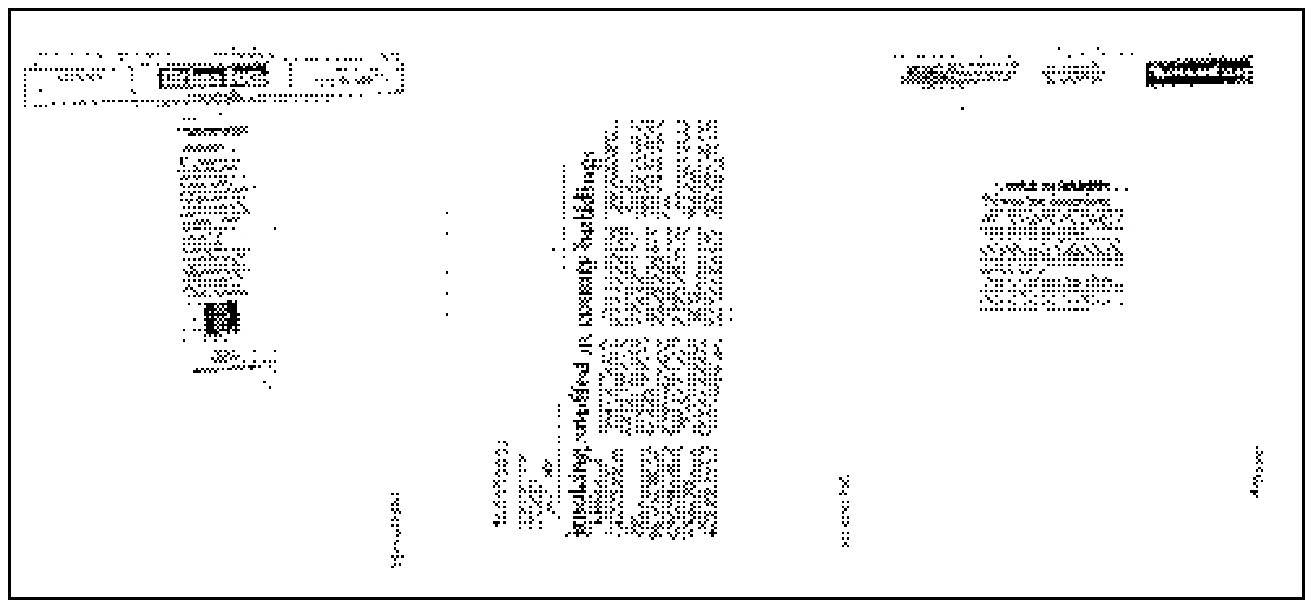}
		\hfill  \includegraphics[height = 0.22\columnwidth]{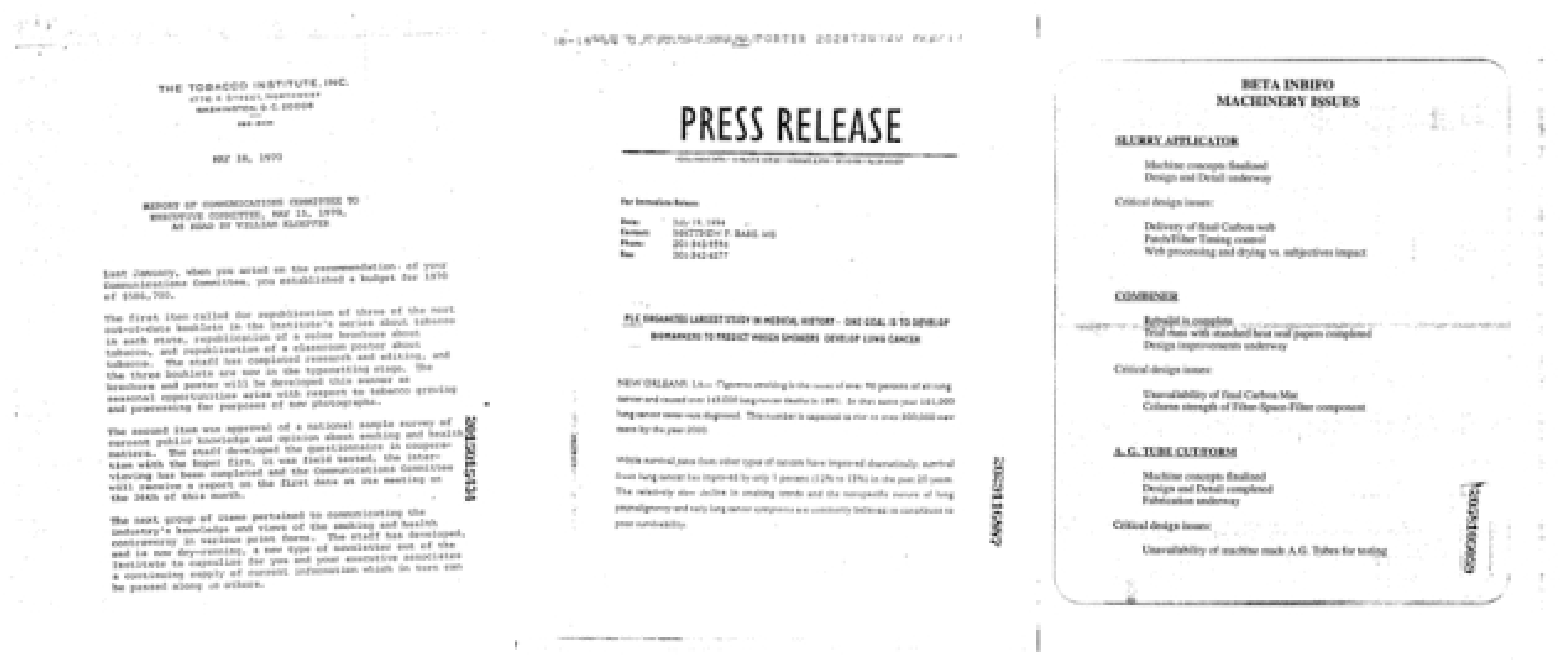}
		\hfill  \includegraphics[height = 0.2\columnwidth]{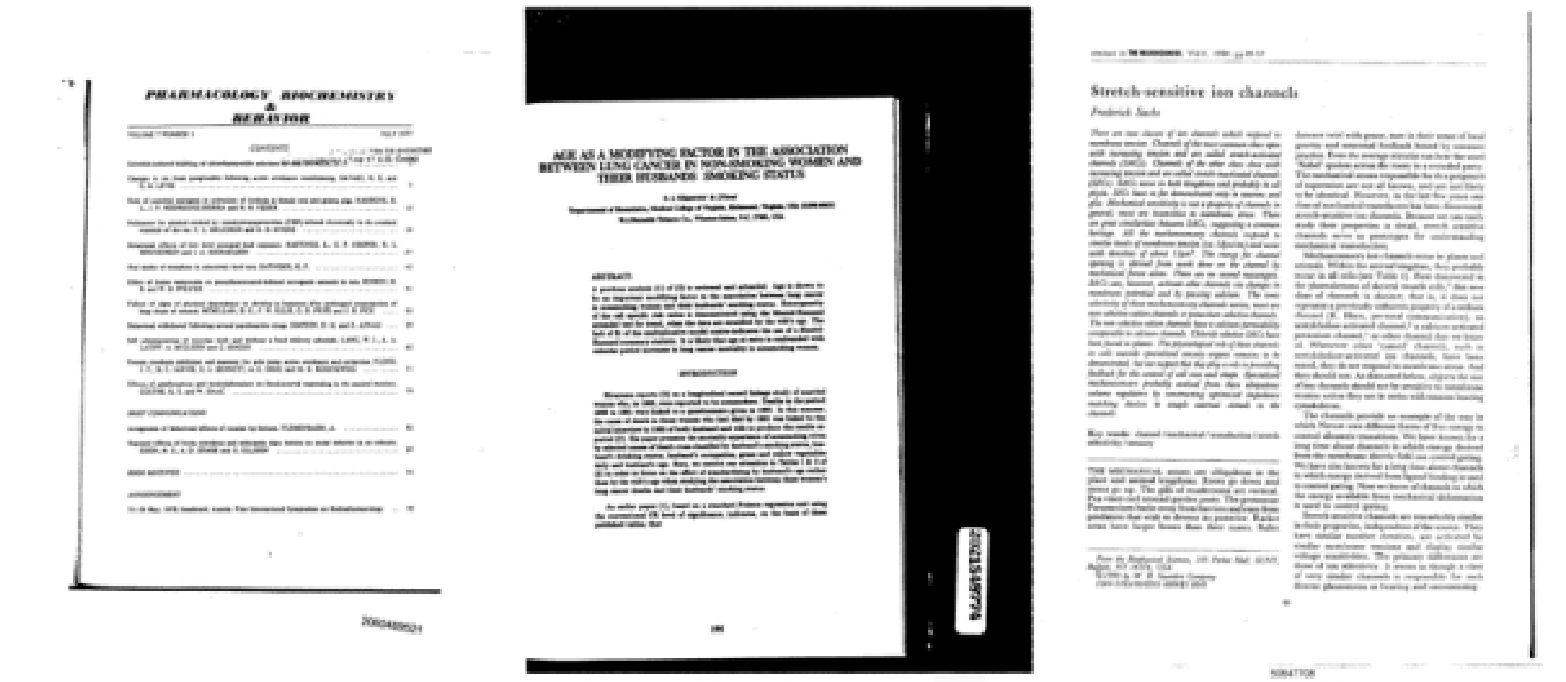}\\
		$(i)$\hspace*{4cm}$(j)$\hspace*{4cm}$(k)$\hspace*{4cm}$(l)$\\
		\vspace*{.25cm}
		\includegraphics[height = 0.22\columnwidth]{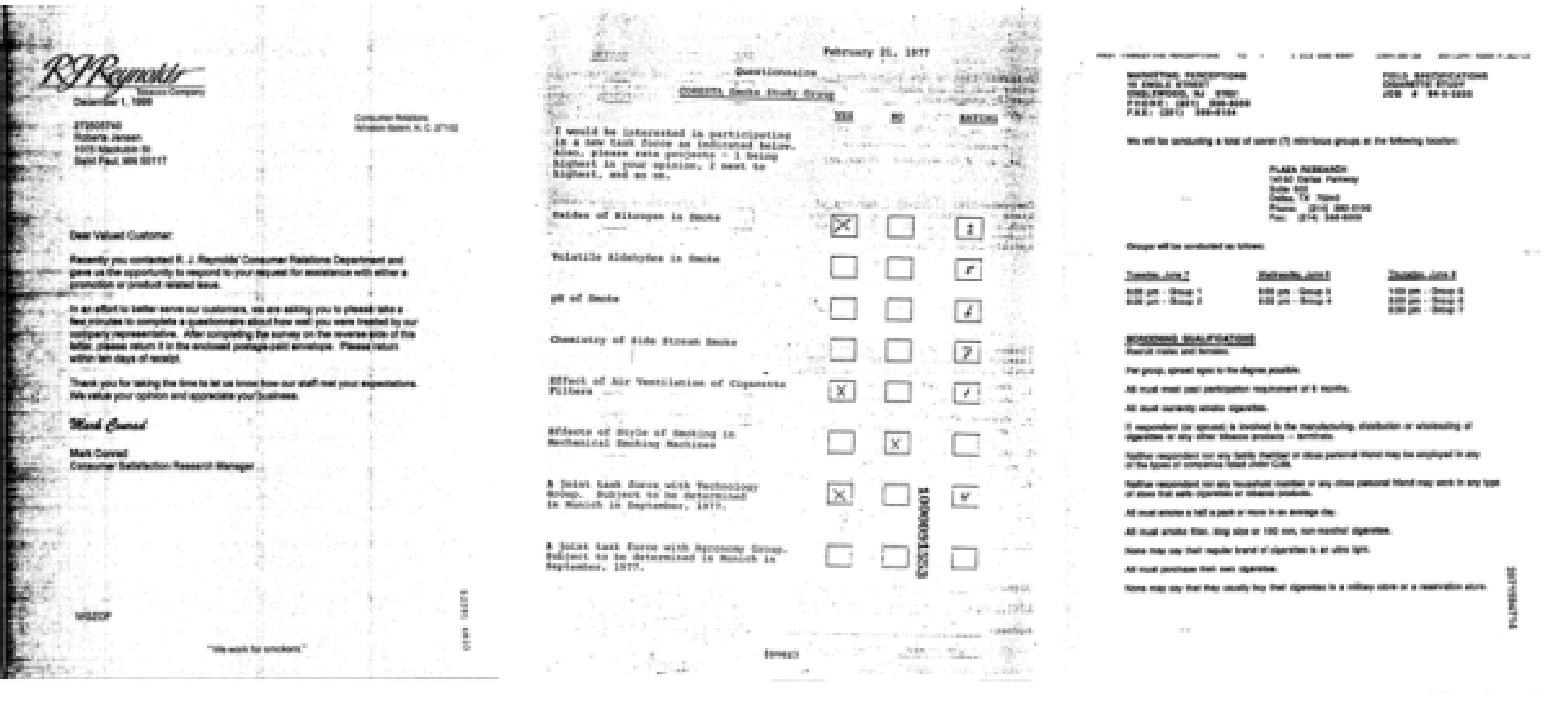}
		\hfill  \includegraphics[height = 0.22\columnwidth]{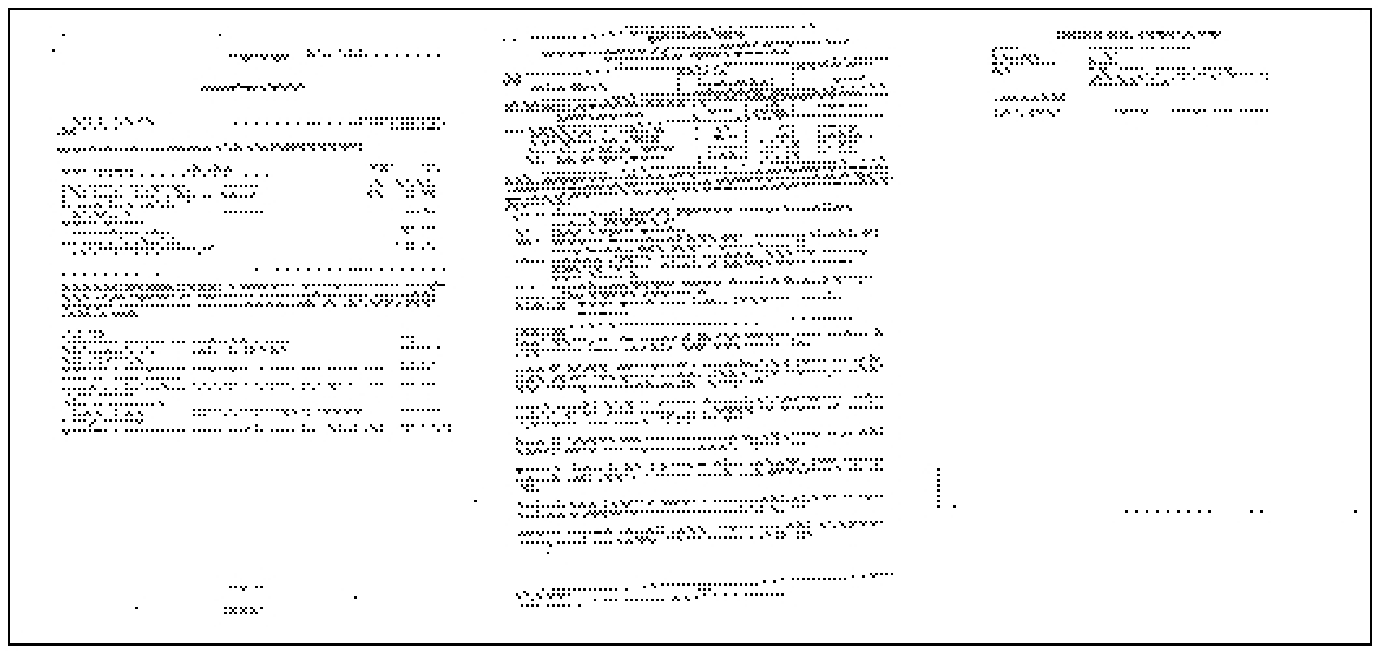}
		\hfill  \includegraphics[height = 0.22\columnwidth]{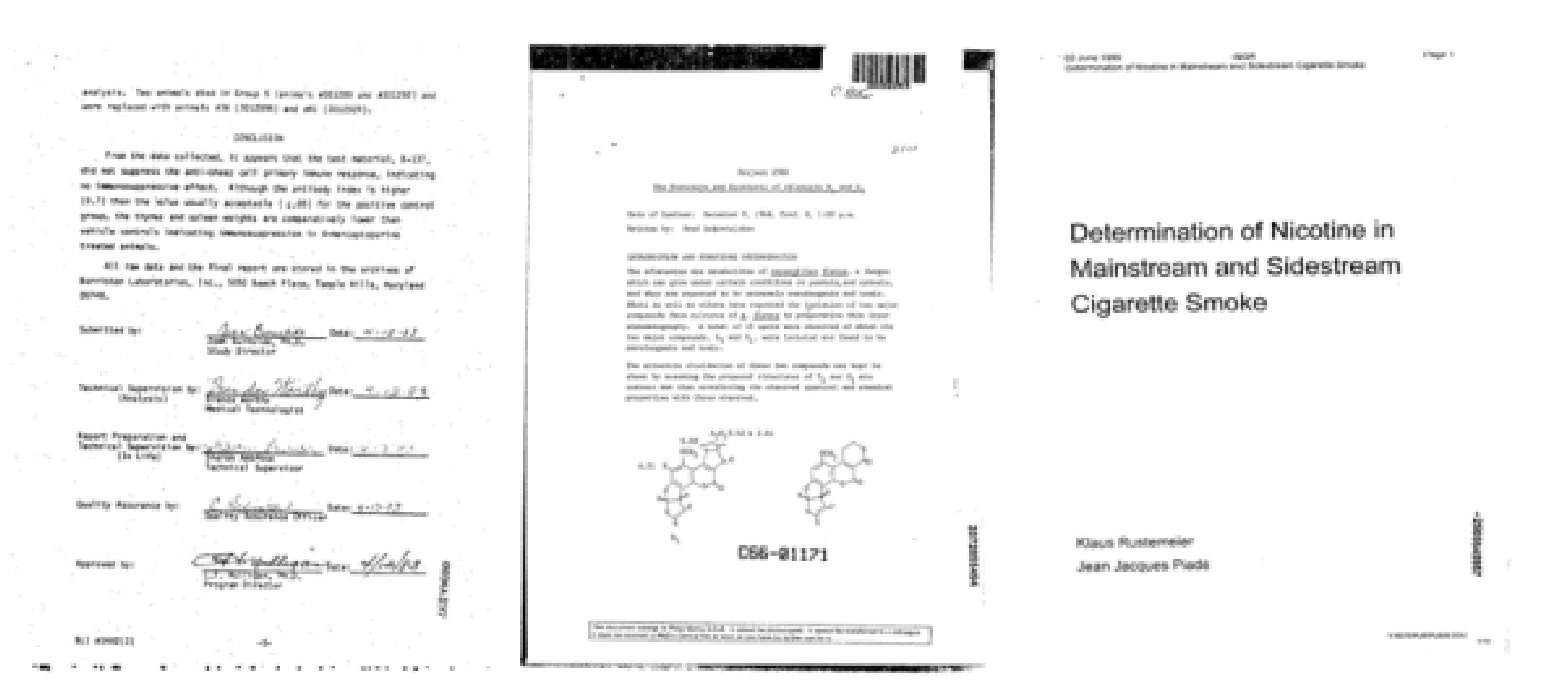}
		\hfill  \includegraphics[height = 0.22\columnwidth]{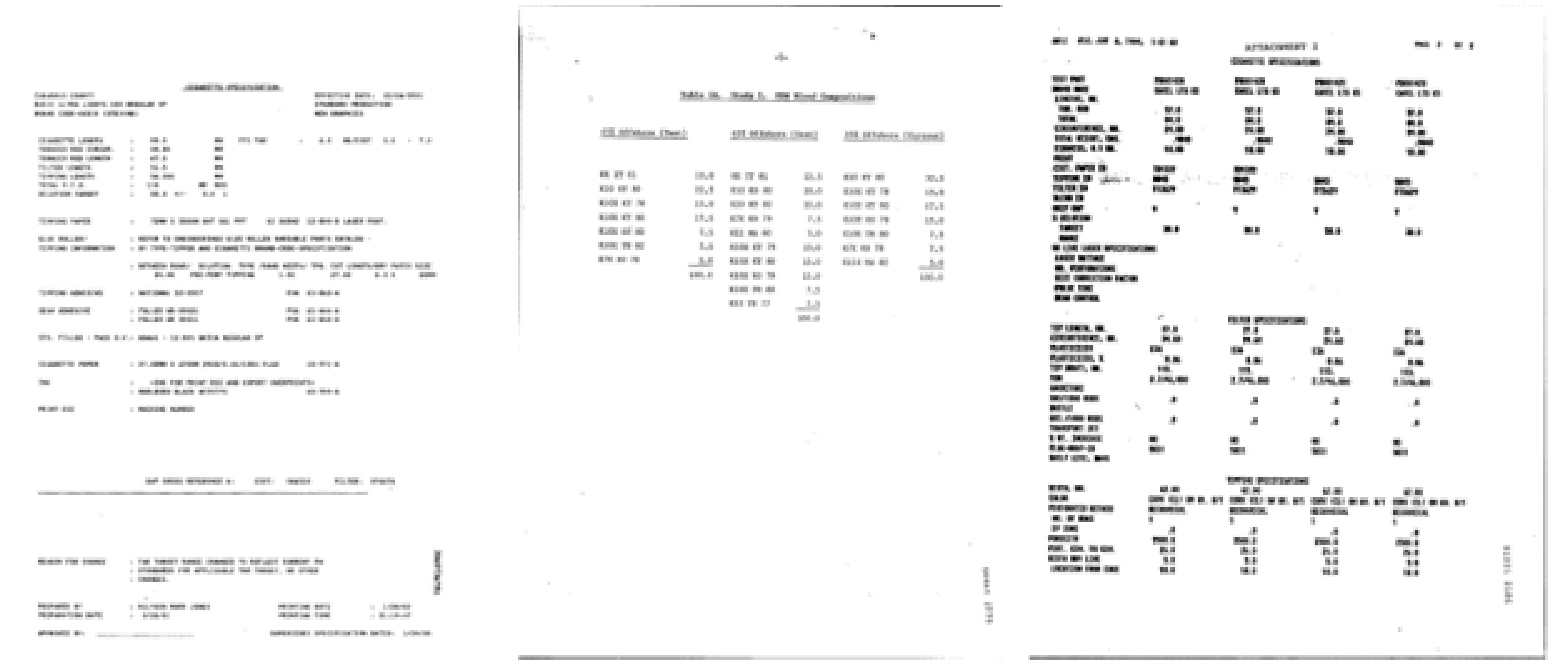}\\
		$(m)$\hspace*{4cm}$(n)$\hspace*{4cm}$(o)$\hspace*{4cm}$(p)$\\
		%  \vspace*{-1cm}
		\caption{Three samples from each of the sixteen classes of the RVL-CDIP database after resizing, preserving only document structure: $(a)$ Letter, $(b)$ Memo, $(c)$ Email, $(d)$ Folder, $(e)$ Form, $(f)$ Handwritten, $(g)$ Invoice, $(h)$ Advertisement, $(i)$ Budget, $(j)$ News, $(k)$ Presentation, $(l)$ Scientific Publication $(m)$ Questionnaire, $(n)$ Resume, $(o)$ Scientific Report $(p)$ Specification.} \label{fig:datasheet}
	\end{center}
	\vspace*{-3mm}
\end{figure*}

With increased usage of digital means of document exchange and the growth in volume of digitized documents, the requirement of an automatic system capable of understanding their logical structures is felt significantly for better management of these documents including their efficient archiving, retrieval, information mining and so on. However, the development of an automatic system for classification of arbitrary document images into their respective true categories is computationally a non-trivial task in contrast to the case of human beings. The complexity of this task is increased due to inter-class similarity and intra-class variability issues in documents. For example, an advertisement may look like a news item or a form may be very dense with items arranged in multiple columns while another form may have only a few items with each item in a distinct horizontal row and the rows are separated wide apart. This is demonstrated by the examples shown in Fig. \ref{fig:datasheet}.

Intuitively, from the perspective of a classifier, documents can be characterized by their \textit{text content} or \textit{structural information}\cite{chen2007}. Document classifiers based on text contents use optical character recognition (OCR) techniques to extract the texts in the document image and thus are susceptible to OCR errors. On the other hand, there are OCR systems which follow a structural analysis approach by first determining the class of the document image based on which an appropriate OCR module is employed \cite{appiani2001}. The effectiveness of such systems hinges on that of the structure classification techniques employed and hence speaks for the utility of structure learning as an important aspect of document understanding.

\subsection{Contribution}
The proliferation and subsequent effectiveness of deep learning techniques in a wide spectrum of tasks in machine learning over the last decade has been spoken of extensively in the literature~\cite{schmidhuber2015deep,lecun2015deep}. Deep neural networks have been used to provide unparalleled results in multiple domains including document image classification. In the present work, deep convolutional neural networks (DCNN) are used for automatically understanding the structural aspects of a document for the purpose of classification. While DCNN based approaches are not new to this area \cite{kang2014, afzal2015, harley2015evaluation}, the present study distinguishes itself by studying the rapid training of effective document region based classifiers. To achieve the same, multiple levels of transfer learning are used. The unique nature of the region based approach to document classification is utilized to make a case for both \textit{inter-domain} and \textit{intra-domain} transfer learning for this problem. Integration of the predictions from region based classifiers are done by performing a thorough study into meta-classification using stacked generalization. Finally, the proposed techniques are validated experimentally by achieving state-of-the-art results on the popular RVL-CDIP dataset containing 400,000 document image samples of 16 different categories.

\section{Related Work}
%Designing of a high-performance document classifier is challenging due to a number of reasons including the large variety of documents in any individual document class \cite{chen2007}. Often the structures of two documents belonging to two different classes are so similar that their correct classification is very difficult. Experimentation of the proposed classification approach has been done on Tobacco litigation dataset \cite{lewis2006} and a few samples from the same have been shown in Fig. 1.

Several studies on automatic document classification have been made in the recent past. A related survey work can be found in \cite{chen2007}. An approach to automatic generation of a decision tree for logical labeling of business letters was proposed in \cite{dengel1993}. A 3-step approach based on certain GTree was studied in \cite{dengel1994} for partitioning a document image into its logical objects. INFOCLAS, an early prototype system  for indexing and classifying printed business letters, was described in \cite{Hoch1994}. In \cite{Junker1997} a machine learning based approach was proposed for automatic discovery of knowledge, that is, the relevant features for document classifiers.

Further, in \cite{cesarini2001} a modified X-Y tree was used to describe a document page and its hierarchical structure provided a fixed length feature vector for a multilayer perceptron (MLP) for classification. In \cite{heroux1998}, structural classifiers in addition to $k$-Nearest Neighbours ($k$NN) and MLP classifiers were used for classification of form document images. In \cite{rabiner1990}, Hidden Markov Models (HMM) were considered to be robust and suitable for handling uncertainties and noise in input document image samples. In \cite{hu2000}, interval coding was considered to compute spatial layout of input document images and the resulting fixed length feature vector was used for its classification based on such an HMM.

In another study \cite{shin2001}, a supervised classifier was trained using  given examples from each underlying class and exploited “visual similarity” of document layout structure for their classification. Similarly \cite{hu2000} also proposed document classification based on the layout similarity. In contrast, \cite{diligenti2003} used a recursive representation of document structure to preserve relationship among its different parts. Also recently, in \cite{gordo2013}, certain multi-scale run length histograms for representation of document images and a generative classifier model for efficient classification were proposed.

\begin{figure}[t]
	\centering
	\includegraphics[height=5cm, width=0.9\columnwidth]{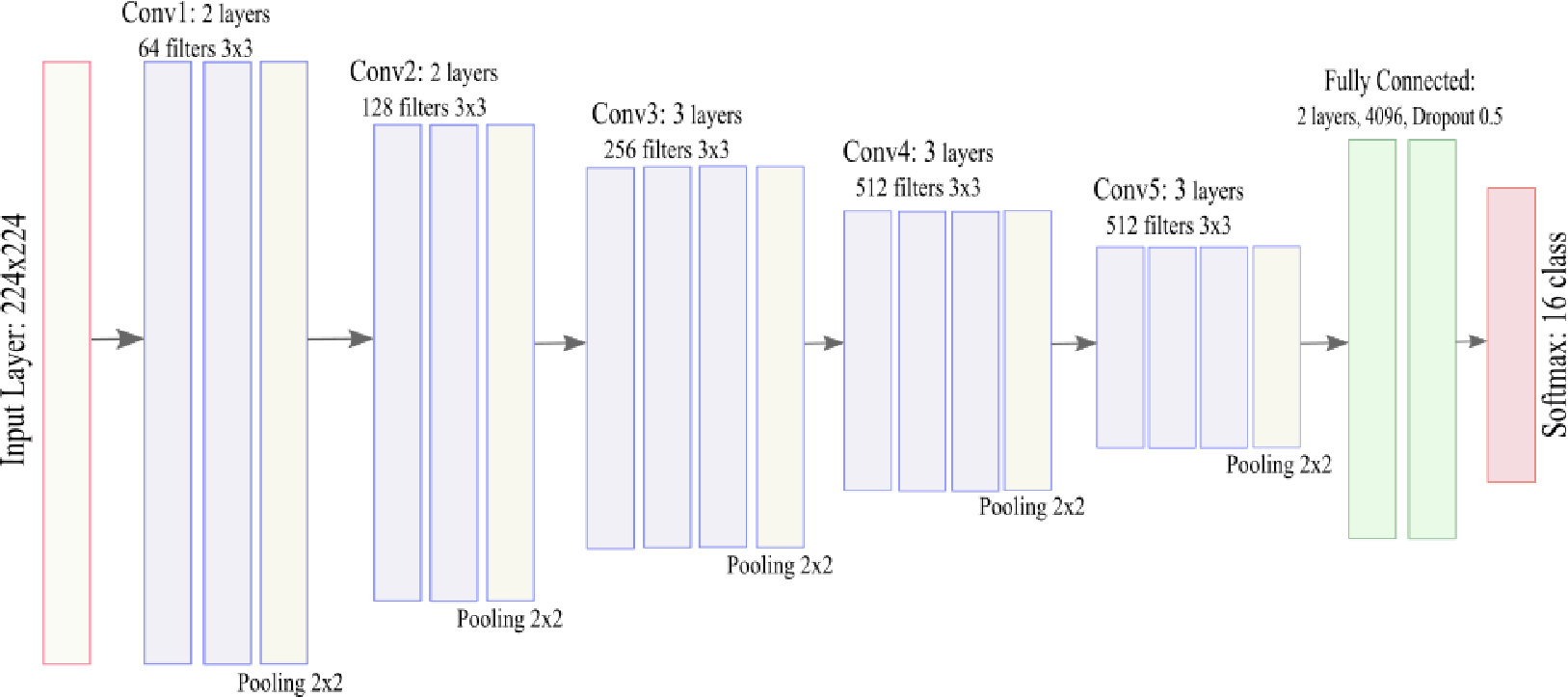}
	\caption{VGG16 Architecture with Softmax layer replaced used as the base model for each classifier} \label{fig:DCNNarch}
\end{figure}

Deep learning techniques which have been popular across multiple domains such as object recognition have also been used in document structure learning tasks. Deep Convolutional Neural Networks (DCNN) are deeper variants of convolutional neural networks \cite{lecun1998} and have recently exhibited their exceptional performance in object recognition tasks \cite{Krizhevsky2012,szegedy2015going}. In \cite{kang2014}, a deep CNN architecture with rectified linear units was trained using dropout \cite{Srivastava2014} for a 10-class document image classification task, popularly known as the Tobacco3482 dataset\cite{Kumar2014}. Later, in \cite{afzal2015}, the concept of Transfer Learning \cite{Pan2010} was used to improve the recognition accuracy on the same standard dataset by using a CNN pre-trained on a ImageNet dataset \cite{deng2009}. Another work in the area which used transfer learning includes \cite{harley2015evaluation} which also introduced region based modelling and introduced the larger 16 class RVL-CDIP dataset. In \cite{roy2016generalized}, a committee of lightweight supervised layerwise trained models on Tobacco3482 achieved decent results without any transfer learning. In \cite{tensmeyer2017analysis}, extensive exploration was done on the variation of components such as architectures, image size, aspect ratio preservation, spatial pyramidal pooling, training set size among others using the AlexNet architecture and the RVL-CDIP and ANDOC datasets. In \cite{csurka2016right}, run-length and fisher vector representations trained on an MLP were compared to AlexNet and GoogLeNet architectures on the RVL-CDIP dataset with deep models showing better performance. Also, in \cite{afzal2017cutting} AlexNet, VGG-16, GoogLeNet and ResNet-50 models were tested using Transfer Learning on the RVL-CDIP and Tobacco3482 datasets. In comparison, \cite{kolsch2017real} concentrated on speed by replacing the fully connected portion of the VGG architecture with extreme learning machines (ELM).

%The goal of our present study is the development of comparatively lightweight architecture consisting of multiple DCNNs each being trained on different parts or whole of document images but still exhibiting acceptable generalization performance making it suitable for low configuration handheld devices and/or in an environment supporting parallel computation. We studied combination of the results of base DCNN classifiers within the methodology of Stacked Generalization by three different schemes. Simulation results have been obtained on a labelled dataset of tobacco litigation documents called the Ryerson Vision Lab Complex Document Information Processing (RVL-CDIP) dataset containing 400,000 document images of 16 different categories and downloaded from \url{http://scs.ryerson.ca/~aharley/rvl-cdip}.
\begin{figure}[]
	\centering
	\includegraphics[height=5cm, width=0.9\columnwidth]{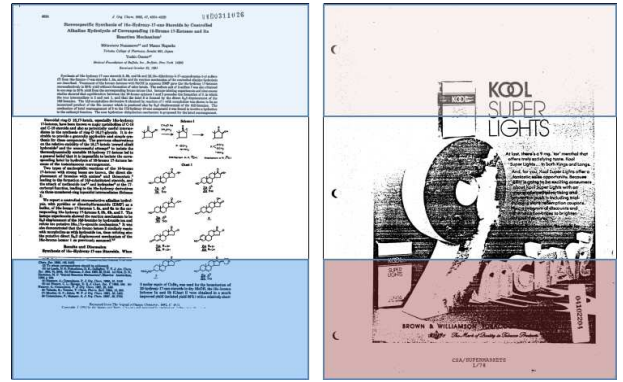}
	\caption{Illustration of characteristic differences between the Header and Footer regions in \textit{Scientific Publication} and \textit{Advertisement} classes} \label{fig:classDif}
\end{figure}

\section{Proposed Method}

\subsection{Deep Convolutional Neural Network Architecure}
DCNNs are currently one of the most popular models for deep learning. DCNNs use many of the ideas used by conventional MLPs such as feed-forward connections, non-linear activations, gradient descent and backpropagation. However, their concept of shared weights and more recently activation functions like the rectified linear unit, $f(x)=max(0,x)$ make them resistant to the vanishing gradient problem and effective as deep learning architectures. Also, their multi-filter weight sharing concept help them discover robust features in input spaces like images and audio data making them powerful supervised classifiers.\cite{Krizhevsky2012,sainath2013deep} As seen earlier, they have found use in the task of document image classification utilizing images containing just the structural framework of a document without much, if any, content being legible.

From \cite{tensmeyer2017analysis, csurka2016right, afzal2017cutting}, it was shown that the VGG16 model \cite{simonyan2014very} performs better on a document classification task than other DCNN models. This information is utilized by us to select the VGG16 model for our base classifier model for this task. The general architecture is shown in Fig. \ref{fig:DCNNarch} with the Adam Optimizer \cite{kingma2014adam} being used for training along with a learning rate decay tuned based on the accuracy on the validation set. The initial weights used are transferred from a model trained on the \textit{ImageNet} object recognition dataset.

\subsection{Region based modelling for Document Classification}
A combination of holistic and region based modelling for document image classification was introduced in \cite{harley2015evaluation}. The general idea consists of training multiple machine learning models to capture influences of holistic as well as region-specific visual cues of various document classes. For example, a \textit{scientific publication} or a \textit{letter} contains distinctively informative header sections than, say, an \textit{advertisement} as seen in Fig \ref{fig:classDif}. In contrast, the central region of a \textit{memo} can be very different from that of a \textit{resume}. This supports the fact that a modelling technique governed by combination of features from multiple document image regions can be effective in their classification. As in \cite{harley2015evaluation} and \cite{roy2016generalized}, the various sections used for our purposes in this work include \textit{Holistic (entire image), Header, Footer, Left Body} and \textit{Right Body}.

\begin{figure}[t]
	\centering
	\includegraphics[height=8.5cm, width=\columnwidth]{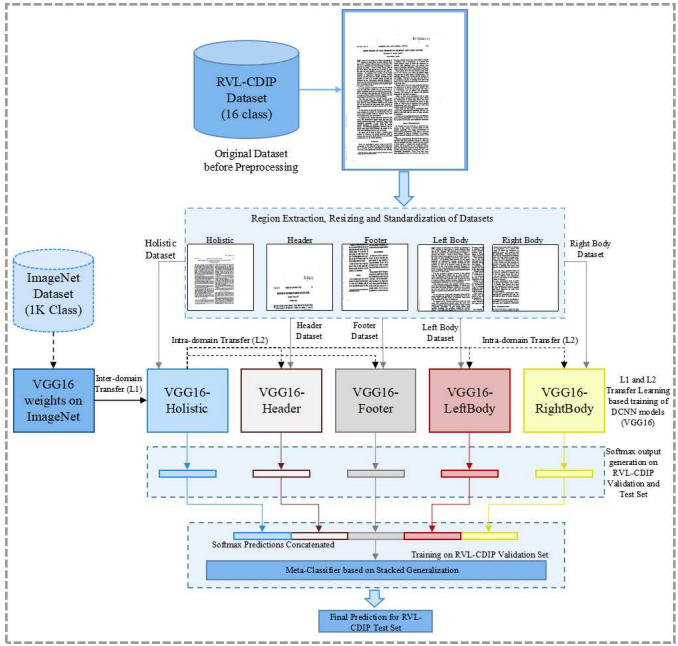}
	\caption{Flowchart of Proposed Model for Document Image Classification with L1 and L2 Transfer Learning} \label{fig:flowchart}
\end{figure}

\subsection{Intra-Domain Transfer Learning of Region DCNNs}
Transfer Learning involves the transfer of experience obtained by a machine learning model in one domain into another related domain \cite{Pan2010}. While document classification and object classification apparently seem like divergent domains, architectures trained on the 1000 class ImageNet dataset have proven to function as \textit{generalized feature extractors}. Region based training of document recognizers on an architecture such as the VGG architectures creates an infeasible learning problem due to the sheer training time of such a model. For example, each VGG16 model contains more than 130 million trainable parameters. This makes training from scratch (VGG16 on ImageNet weights) on each region-based model computationally prohibitive. However, the nature of training region based models creates a unique scenario that can be used to bypass this predicament.

In this work, a VGG16 model, trained on ImageNet, is used as initial weights of our holistic model thus constituting an initial level-1 (L1) transfer of weights. The L1 transfer, of course, originates from a different domain and is the regular inter-domain form of transfer learning. However, the holistic model trained on whole images of the RVL-CDIP dataset can be thought of as a \textit{generalized document feature extractor}. The training sets for the region based models, although containing images of document regions and at a different scale, are still essentially images of documents. Thus, this concept is utilized by setting up another level (L2) of transfer learning in which the region based models are initialized with weights from the holistic model instead of the original VGG16 model.

The DCNN models trained are illustrated in Fig.\ref{fig:flowchart} and can be summarized as follows:

\begin{itemize}
\item[--] \textbf{VGG16-Holistic}: Trained with whole images from the RVL-CDIP dataset. The initial weights transferred (L1) from the VGG16 model trained on the ImageNet dataset.
 	
\item[--] \textbf{VGG16-Header/Footer/LeftBody/RightBody}: Trained with the \textit{Header/Footer/Left Body/Right Body} section of RVL-CDIP images. The initial weights are transferred (L2) from the fully trained VGG16-Holistic model.
\end{itemize}

The L2 transfer is shown to demonstratively provide dividends by needing only a few iterations of fine tuning to provide excellent results. In this work, each region based model was only fine tuned for 4 iterations after L2 transfer with merely a decaying learning rate. In contrast, the VGG16-Holistic model took 25 iterations after an L1 transfer to provide equivalent convergence. This demonstrates a significant benefit for the intra-domain transfer learning approach.

\subsection{Stacked Generalization Schemes for Model Aggregation}
The strategy of stacking or stacked generalization was introduced by Wolpert \cite{wolpert1992}. It is an ensemble learning technique for machine learning models designed to produce a reduced generalization error than the same of its individual base models. The idea has been used to a limited extent in the literature to combine DCNN models. It was shown in \cite{roy2016generalized} that stacked generalization performs significantly better than the simple winner-takes-all approaches.
%This phase involves the following steps.	
%\begin{itemize}
%	\item Train each of the base DCNN model using the training samples.
%	\item Test the trained DCNN models on the validation set of samples.
%	\item Concatenate the sets of outputs of all the trained base DCNN models corresponding to a validation sample and arrange the same in a row.
%	\item Assign the true class label to each such row (number of rows equal to the number of validation samples) forming the training set for the meta classifier.
%	\item Train the meta classifier using the training samples formed as in the above.
%	\item Obtain the prediction results by the meta classifier on novel test samples which was not used before during the past phases of the learning session.
%\end{itemize}
Stacked generalization works by training a meta-classifier on predictions of the base classifiers on a hold-out dataset to \textit{learn} the final set of predictions. In the present study, the region-based DCNN models were considered to be the base models for stacked generalization. The concatenation of the base class softmax predictions on the validation set for each base model are used to train the meta-classifier while the same on the test sets are used to get the final prediction.

Generalizing, let $Q^j_i$ be the $c$-dimensional vector consisting of predicted probability values of the $i^{th}$ classifier for the $j^{th}$ data sample corresponding to all of the underlying $c$ classes. If $f(.)$ be a meta-classifier and $n$ is the number of base classifier models, then a meta-classifier can be said to learn the mapping $ f: \mathbb{R}^{c\times n} \rightarrow \mathbb{R}$

The domain of $f$, that is, the feature space for the meta-classifier is basically just an aggregate space of the softmax outputs for each base DCNN model. Infact, the meta-feature for a sample $j$ is simply $Q^j_1 \wedge Q^j_2 \wedge ... \wedge Q^j_n$ where $\wedge$ represents concatenation of the prediction vectors from the base classifiers.

For this work, $c=16$ and $n=5$. The three schemes studied here for generation of the meta classifier are detailed below.

The meta-classifiers used in the proposed method include:
\begin{enumerate}
	\item \textbf{Linear Regression}: Involves learning a linear mapping of the input to predict the output.
	\item \textbf{Ridge Regression}: Linear regression with \textit{l2}-regularization.
	\item \textbf{K-Nearest Neighbours (kNN)}: Instance space learning algorithm which classifies a point by the majority of the labels of its k nearest neighbours. Values of k tested were 32, 64 and 128.
	\item \textbf{Support Vector Machine (SVM)}: SVMs are popular supervised learning models which are maximum margin separators. Here, an SVM with an RBF kernel was used.
	\item \textbf{Bootstrap Aggregating (with SVM)}: Also known as bagging, it is a meta-classifier which works by sampling subsets (bags) of the input data and training multiple base classifiers (SVM in this case) and aggregating the results. Here, 30 bootstrap samples each with 7500 training examples were used.
	\item \textbf{Extreme Learning Machine (ELM)}: ELMs (described simplistically) involve randomly assigning weights in the hidden units and train extremely fast compared to traditional neural networks. A hidden layer of 100 units was used in our experiments.
	\item \textbf{Multilayer Neural Network (MLNN)}: A 3-layer fully connected neural network (256-256-16) with ReLU activation units and heavy dropout rate of 0.75 to control overfitting between layers trained using Adam optimizer.
\end{enumerate}

In case of regression techniques used under this scheme, a simple \textit{argmax(.)} was used on the output in a multivariate regression problem. The meta-classifiers were chosen to be as lightweight as possible, so as to not add further to the computational cost of training the system.

%\subsubsection{Scheme III - Combined prediction vector classification}
%Scheme III uses data organized as in Scheme II. However, in contrast to Scheme II, a multiclass classification problem with a number of choices as classification algorithms for the meta-classifier instead of regression. The classification techniques studied here under this scheme are k-nearest neighbour (kNN), Support Vector Machine (SVM) with RBF kernel and a fully connected MLP network.

%Here, $argmax(.)$ of the regression or classification outputs of the meta classifier combining the ensemble is reported as the predicted class.

\section{Experiments and Results}

\begin{table}[]
	\centering
	\caption{Comparison of accuracy \% between Harley et al. and Proposed Method (rounded off to 1 decimal place) on the RVL-CDIP Test Set}
	\label{Tab:RegionAccs}
	\begin{tabular}{|c|c|c|}
		\hline
		\textbf{DCNN Model} & \textbf{Harley et al.} & \textbf{Proposed Work} \\
		\hline \hline
		Holistic            & 89.8                   & 91.1                  \\
		Header              & 84.9                   & 86.0                  \\
		Footer              & 79.4                   & 81.2                  \\
		Left Body           & 82.7                   & 85.2                  \\
		Right Body          & 79.5                   & 82.2                  \\
		\hline
		Ensemble of Models  & 89.3                   & \textbf{92.2} \\
		\hline
	\end{tabular}
\end{table}

\begin{table}[]
	\centering
	\caption{Comparison of accuracies on RVL-CDIP of best models from different techniques}
	\label{Tab:FinalComp}
	\begin{tabularx}{\columnwidth}{|c|c|m{0.444\columnwidth}|}
		\hline
		\textbf{Publication} & \textbf{Accuracy} & \textbf{\hspace{0.16\columnwidth}Comments}\\
		\hline \hline
		Harley et al.\cite{harley2015evaluation}        & 89.80\%            & {Document section-based models with weight transfer from Alexnet with max voting ensemble \centering}\\
		\hline
		Tensmeyer et al.\cite{tensmeyer2017analysis}     & 89.31\%           & Uses AlexNet architecture with spatial pyramidal pooling, without transfer learning\\
		\hline		
		Tensmeyer et al.\cite{tensmeyer2017analysis}    & 90.94\%           & Same architecture as above with images resized to $384\times 384$ and aspect ratio preservation\\
		\hline
		Csurka et al. \cite{csurka2016right}        & 90.70\%           & The GoogLeNet architecture was used with ImageNet based transfer learning\\
		\hline
		Afzal et al. \cite{afzal2017cutting}        & 90.97\%           & Weights transfer from AlexNet, VGG-16, GoogLeNet and ResNet-50\\
		\hline		
		Proposed Work         & \textbf{91.11\%}           & Trained on full document images with weights transfer from VGG-16 trained on Imagenet\\
		\hline
		Proposed Work         & \textbf{92.21\%}           & MLNN based stacking of holistic \& region-based models with inter and intra-domain weights transfer\\
		\hline
	\end{tabularx}
\end{table}

\begin{figure}[t]
	\centering
	\fbox{\includegraphics[height=5cm, width=0.97\columnwidth]{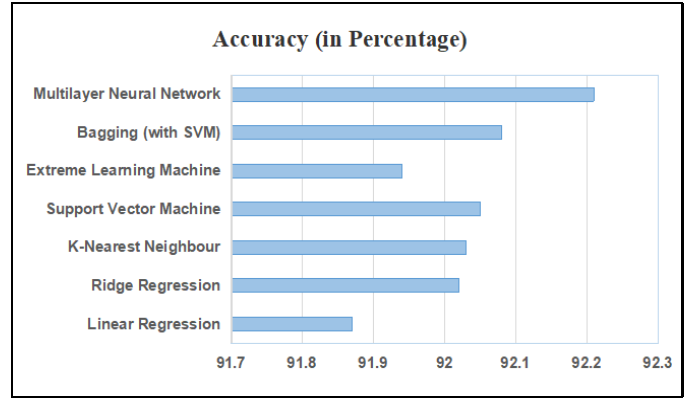}}
	\caption{Comparison of Accuracies for different meta-classifiers in Stacked Generalization} \label{fig:stacking}
\end{figure}

\subsection{Dataset}
The developed model was tested on a subset of the IIT-CDIP Test Collection known as the RVL-CDIP dataset. The dataset consists of scanned grayscale images of documents from lawsuits against American Tobacco companies and is segregated into 16 categories or classes. A sample of the dataset can be seen in Fig. \ref{fig:datasheet}. The dataset is subdivided into Training, Validation and Test Sets each containing 320000, 40000 and 40000 images respectively.

\subsection{Preprocessing}
The preprocessing for holistic and section-based datasets were only marginally different. The images for the \textit{Holistic} dataset were initially resized to $224 \times 224$. For the region-based images, the regions were extracted exactly as in \cite{harley2015evaluation}, for the ease of comparison. The extracted regions were further resized to $224 \times 224$. Following the resizing, all datasets were standardized and the single image channel was duplicated to 3 channels for VGG16 compatibility.

\subsection{Evaluation}
It is seen in Table \ref{Tab:FinalComp} that the VGG16-Holistic model with L1 transfer by itself achieves a state of the art accuracy on the RVL-CDIP dataset with Adam optimizer and gradual learning rate decay based on the validation set accuracy. Of course, the stacked generalized ensemble performs even better setting a benchmark at 92.21\%. However, if the region models were trained simply with L1 transfer, the gain could be argued to be too less given the computational cost of training multiple large models (each model having ~130 million parameters).

As mentioned earlier, while the holistic model took about 25 iterations after the L1 transfer to converge, the VGG16-\textit{Region} models took only 4 iterations after L2 transfer to show comparable results. The performance can be further evaluated by referring to Table \ref{Tab:RegionAccs}. Every region based model in this work after only a few iterations of training (or fine tuning) easily surpasses the slightly smaller AlexNet models trained in \cite{harley2015evaluation} by L1 type transfer learning. The effectiveness of the L2 transfer significantly reduces training time for the region based models and eases the computational cost of training multiple models, while preserving benefits characterized by the gain in accuracy.

Finally, extensive experiments into meta-classifiers for stacked generalization demonstrate that fully connected MLNNs, otherwise known as MLPs, are the clear favorite as the second level meta-classifier. The second best performance, perhaps unsurprisingly, is by Bagging since it is itself an ensemble technique utilizing the second best individual meta-classifier in this work, that is, SVMs. Fig. \ref{fig:stacking} demonstrates the comparative predictive performance of the various meta-classifiers.

%Another perhaps interesting statistic, albeit with limited usefulness in a scenario for classification, is that the top-3 accuracy for our final model is about 97.5%.

\section{Conclusion}
In this work, rapid training of region based DCNNs has been explored with multiple levels of transfer learning. The similarity between region based and the holistic input spaces facilitates the usage of intra-domain transfer learning in addition to the general inter-domain variant. This allows fast convergence and feasibility of region based models in what would otherwise be an unnecessarily unwieldy machine learning problem. Also, a thorough experimentation of stacked generalization has been performed with multiple meta-classification algorithms. Finally, a state-of-the-art result was not only set on the holistic model at 91.11\% but also on the final stacked generalized model at 92.21\% accuracy on the RVL-CDIP dataset, representing a significant gain over the existing methods. In the future, faster but less accurate models such as \cite{afzal2017cutting} could be explored as with region based intra-domain transfer for building effective ensemble models.

\bibliographystyle{IEEEtran}
\bibliography{refs}

\end{document}